# Reducing catastrophic forgetting of incremental learning in the absence of rehearsal memory with task-specific token

Young Jo Choi, M.S[a], Min Kyoon Yoo, B.S[a], Yu Rang Park, Ph. D[a*]

[a]Department of Biomedical Systems Informatics, Yonsei University College of Medicine, Seoul, Republic of Korea.

* Yu Rang Park, PhD

Department of Biomedical Systems Informatics, Yonsei University College of Medicine,

50-1 Yonsei-ro, Seodaemun-gu, Seoul 03722, Korea

Tel: 82-2-2228-2493, E-mail: yurangpark@yuhs.ac

**Abstract**

Deep learning models generally display catastrophic forgetting when learning new data continuously. Many incremental learning approaches address this problem by reusing data from previous tasks while learning new tasks. However, the direct access to past data generates privacy and security concerns. To address these issues, we present a novel method that preserves previous knowledge without storing previous data. This method is inspired by the architecture of a vision transformer and employs a unique token capable of encapsulating the compressed knowledge of each task. This approach generates task-specific embeddings by directing attention differently based on the task associated with the data, thereby effectively mimicking the impact of having multiple models through tokens. Our method incorporates a distillation process that ensures efficient interactions even after multiple additional learning steps, thereby optimizing the model against forgetting. We measured the performance of our model in terms of accuracy and backward transfer using a benchmark dataset for different task-incremental learning scenarios. Our results demonstrate the superiority of our approach, which achieved the highest accuracy and lowest backward transfer among the compared methods. In addition to presenting a new model, our approach lays the foundation for various extensions within the spectrum of vision-transformer architectures.

**Keywords**: Continuous learning, incremental learning, task-incremental learning, catastrophic forgetting, security.

## 1. Introduction

Deep learning algorithms replicate the human learning process by constructing neural networks. However, unlike sequential knowledge acquisition in humans, deep learning algorithms require training on all data simultaneously to achieve optimal results. During the learning of sequential tasks, knowledge of the previous task is often overwritten when new classes or domains are learned. This is called 'catastrophic forgetting.' To mitigate this issue, several research methodologies, known as 'continual learning' and 'incremental learning,' have been proposed (De Lange et al., 2022; D.-W. Zhou et al., 2023). Incremental learning methods aim to balance stability (i.e., effectively remembering past knowledge) with plasticity (i.e., learning new knowledge effectively) (Grossberg, 2013). Depending on the characteristics of the data used for additional learning, these approaches can be categorized into three types: domain incremental learning (domain-IL), task-incremental learning (task-IL), and class-incremental learning (class-IL) (van de Ven et al., 2022). In the case of domain IL, although the new incoming data had identical output classes, their distributions differed. The common aspect of task-IL and class-IL is that they learn more about the data with new output classes. However, the difference is that information regarding the task being performed is provided in task-IL and not in class-IL during the inference phase.

Recent approaches to incremental learning have focused on reusing past data within limited memory frameworks, commonly referred to as rehearsal memories or exemplars (Rebuffi et al., 2017; Robins, 1993; ROBINS, 1995). During the learning process in such an approach, data from a new task are used to acquire new knowledge, and rehearsal memory is used to prevent forgetting previous knowledge. However, this approach poses potential security and privacy risks, particularly when using highly sensitive data (De Lange et al., 2022). The use of rehearsal memory requires access to and the potential export of raw data from previous tasks. Such problems become fatal in scenarios involving confidential information, such as medical records, personal data, or business-sensitive materials, where external data transfer is generally restricted (Thapa & Camtepe, 2021). Although previous studies demonstrated incremental learning algorithms without rehearsal memory (Huang et al., 2023; Petit et al., 2023; Zhu et al., 2021), they commonly display the problem of forgetting knowledge from previous tasks.

To address these challenges, we developed a novel architecture to minimize catastrophic forgetting using only the data from the current task without relying on rehearsal memory. That is, this model can generally be applied regardless of the security constraints. Our approach combines the characteristics of a vision transformer (ViT) (Dosovitskiy et al., 2021) with those of distillation techniques (Hinton et al., 2015). Inspired by the capability of ViT to condense information into a single-class token for classification, we assigned a special token to each task (Douillard et al., 2022). This enabled the attention mechanism to perform differently for each task. Moreover, we optimized the distillation technique to efficiently incorporate the compressed information from previous tasks in the absence of rehearsal memory. Using this approach, we eliminated the risk of storing data while efficiently transferring previous knowledge.

The objectives of this research are as follows: 1) to develop an incremental learning model that eliminates rehearsal memory and reduces catastrophic forgetting and 2) to achieve a task-incremental learning performance higher than that of existing non-rehearsal-based methods.

## 2. Related Works

### 2.1 Incremental Learning

Incremental learning approaches can be categorized into three types: domain incremental learning (domain-IL), task-incremental learning (task-IL), and class-incremental learning (class-IL) (van de Ven et al., 2022). The units of data used for training are referred to as tasks, steps, contexts, or episodes, and these categories are further organized based on the task structure. For domain IL, although the new incoming data have identical output classes, their distribution differs. A common aspect of task-IL and class-IL is that they learn more about the data with new output classes. However, the difference is that information regarding the task being performed is provided in task-IL and not in class-IL during the inference phase.

Many of these methods have been used to balance stability and plasticity. Data-replay methods collectively refer to attempts to access old data. This is typically achieved by storing raw images directly using rehearsal memory. However, some methods also store features extracted from models (Iscen et al., 2020) or train generative models to learn new tasks (Xiang et al., 2019). Regularization-based methods use old data to adjust the direction of weight optimization such that they do not lose information about the old data (Kirkpatrick et al., 2017; Zhao et al., 2020). Dynamic networks organize a network specifically for a task and expand and integrate it as the number of tasks increases (Douillard et al., 2022; Wang et al., 2022; Yan et al., 2021; D.-W. Zhou et al., 2022). Distillation (Xiang et al., 2019) is a method of transferring knowledge by adjusting the loss term to produce outputs for old data, similar to the previous task's model, while learning new data (Douillard et al., 2020; P. Zhou et al., 2020). These are the most common methods used in incremental learning, and one or more can be combined. When learning a new task, most methods use new data to acquire new knowledge, and old data to prevent forgetting. These approaches discuss the best way to organize the rehearsal memory to represent past data (Bang et al., 2021; Rebuffi et al., 2017).

Methods that do not use rehearsal memory have recently been studied. PASS (Zhu et al., 2021) uses an alternative to rehearsal memory by extracting prototypes for each class and adding noise. Furthermore, distillation and data augmentation through self-supervised learning improved the performance. FeTrIL (Petit et al., 2023) trains only the feature extractor of the first task model and then generates pseudo-features by applying a geometric translation to each task's data. Geometric translation is the process of adding the difference between the centroid of the features of the old and new data. The pseudo-feature is then trained using a neural network or support vector machine. TCIL (Huang et al., 2023) was originally a rehearsal-based method; however, it showed high performance even in the absence of rehearsal memory. TCIL uses multi-level distillation (P. Zhou et al., 2020) for logit and features, and rescoring (Zhao et al., 2020) such that the weight norm of the part for new data is on a scale similar to the weight norm for old data. TCIL also uses convolutional neural network (CNN) attention (Woo et al., 2018) to refine and combine features for classification. TCIL's rehearsal-free approach removes the distillation of features from the multi-level distillation. Existing non-rehearsal methods do not directly use old data but create similar synthetic data and use them to learn new tasks. However, these methods are unstable and do not effectively prevent catastrophic forgetting when the number of steps increases. Our experiments show this tendency and that catastrophic forgetting occurs much less frequently in our model than in other non-rehearsal methods. Furthermore, while other models are designed for incremental task growth from the initial task, in the Methods section, we show that our model has a flexible structure that allows for additional knowledge accumulation on an already trained model as long as the model is of the vision transformer series.

## 2.2 Vision Transformer

Transformer (Vaswani et al., 2017) was first developed as a model for machine translation using self-attention. Transformer consists of an encoder and a decoder composed of several layers, including self-attention, and does not use a recurrent neural network layer. The encoder, which is responsible for embedding the input data, evolved into BERT (Devlin et al., 2019), which can effectively perform downstream tasks. Based on the effective performance of Transformer and BERT in the NLP field, vision transformer (ViT) (Dosovitskiy et al., 2021) was developed in the vision field, which can effectively classify images using self-attention without a convolutional layer. ViT partitions image data into patches and uses them as tokens. After tokenization, such as BERT's class token, a learnable parameter that contains key information for image classification is added to the front of the patch token sequence. Using self-attention, each token employed as a query vector is transformed into an embedding that reconstructs the entire image. The embedding formed from the class token is directly connected to the multi-layer perceptron (MLP) head to execute the classification task. Consequently, the embedding of the class token accumulates information critical for classification during the learning process. Therefore, a class token can be perceived as directing the overall orientation of self-attention, specifically for classification.

In many cases, ViT, which exhibits lower generalization efficiency on smaller datasets than CNNs, is commonly subjected to pretraining on voluminous datasets. This is followed by the fine-tuning of smaller, specific datasets to enhance the performance of the targeted downstream tasks. To overcome this limitation, several studies have been conducted using basic architectures, such as DeiT (Touvron et al., 2021), ConViT (d'Ascoli et al., 2022), Swin (Liu et al., 2021), CvT (Wu et al., 2021), and CCT (Hassani et al., 2022).

## 2.3 Incremental Learning in Vision Transformer

The use of transformers for incremental learning has also been investigated. DyTox (Douillard et al., 2022) sets aside class tokens from ViT for each task to perform task-specific classification. DyTox consists of an encoder and decoder, each of which is composed of self-attention blocks. During the learning process, the encoder is asked to construct embeddings in the same manner for the data from all tasks, and the reconstructed embeddings are attached to class tokens corresponding to each task ID as input to the decoder. Because different tasks have different class tokens, they are called task tokens. Despite having the same decoder parameters, data from different tasks are embedded in different ways because of the role of task tokens, and different classifiers (MLP heads) are attached to these embeddings to perform classification.

Despite its novelty, the DyTox is based on rehearsal memory. To avoid direct data transfer, we developed a new architecture that eliminates the need for rehearsal memory. Our inspiration originates from the concept of task tokens. By involving the task token in the overall attention process, we created a task token for the direction of attention, which effectively eliminated rehearsal memory. In addition, by performing multiple distillations according to the task token, we obtained the effect of receiving information from multiple teacher models for each task and preventing catastrophic forgetting. Because there is no old data (rehearsal memory), distinguishing between different tasks during training is challenging. Consequently, we concentrated on optimizing the performance in task-IL scenarios by obtaining a similar effect as maintaining multiple models separately for each task.

## 3. Methods

The model maintains a single feature extractor and multiple MLP heads, one for each step. As shown in Fig. 1, the single-feature extractor follows the ViT structure [12]. A task token corresponds to each task to control the direction of the feature extraction. When an input image is assigned a task ID, feature extraction is performed. The extracted features are classified using a task-specific MLP head. Our model closely follows the basic structure of the ViT. However, it uses 1) feature-level distillation in the training phase to enable task-specific information accumulation without rehearsal memory and 2) the gated positional self-attention (GPSA) layer from ConViT (d'Ascoli et al., 2022) for incremental learning to work well with small amounts of split data. The model has two types of losses to ensure that it does not forget information from previous tasks while learning data from new tasks.

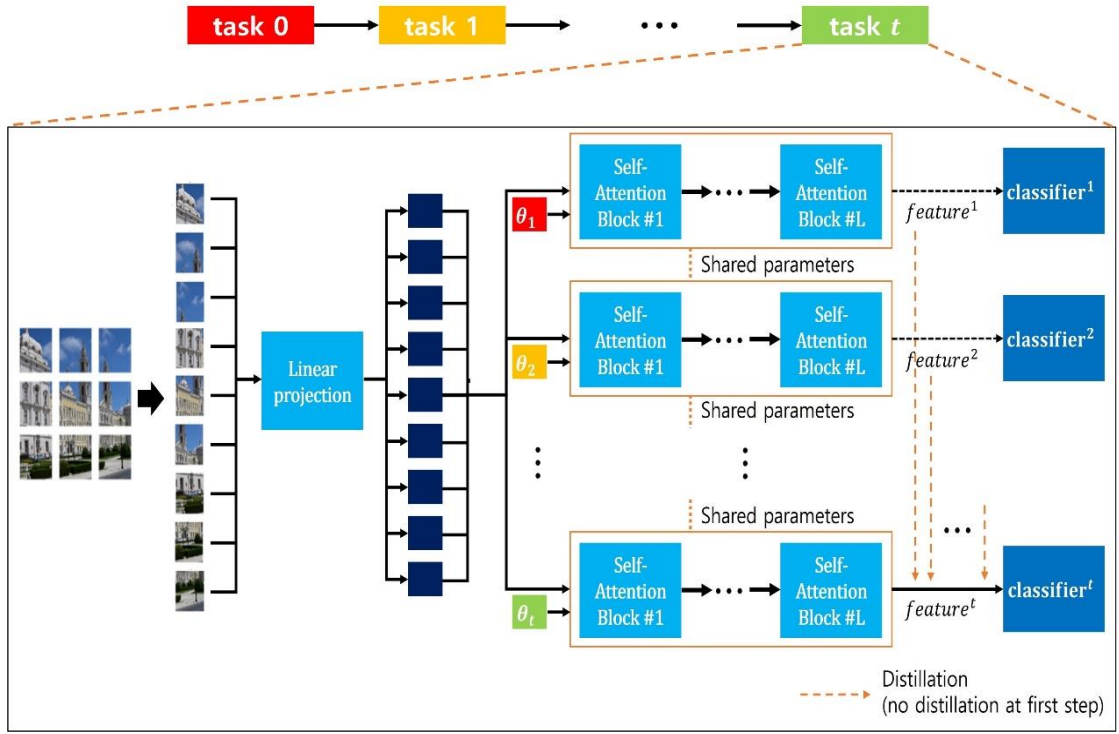

**Fig. 1. Overall model structure.** After patch-by-patch segmentation of the image, a token for each task is prefixed to the patch sequence (such as the class token of a ViT) to learn the classification. In addition, the model from the previous step is used as a teacher to distil past knowledge, and the current model becomes a student. If the task tokens of the previous steps that have already been learned in advance are used, embedding is generated to fit the classification of that task, and the student model distils the teacher model for all the task tokens of the previous step.

## 3.1. Task Token

ViT tokens image $x$ by dividing it into patches. The token is expanded by attaching a class token shared by each batch, which constitutes the input to the self-attention layer. Through self-attention, all the tokens are reconstructed using a query token. The embedding reconstructed with the class token is connected to the MLP head for classification, and the compressed information that distinguishes the classes is collected in the class token. As shown in Fig. 1, we generate a task token for each task, pass it through the self-attention block in conjunction with the tokens to function as a class token in ViT, and compress the classification information of the classes contained in the task. Similar to the ViT, our model's process of learning the classification information of a new task for an image token $x_p \in R^{N \times D}$ tokenized with $N$ patches of dimension $D$ at step $t$ is as follows:

$$z_0 = [\theta_{task}^t; x_p^1 E; x_p^2 E; \cdots, x_p^N E], \quad E \in \mathbb{R}^{(P^2 \cdot C) \times D}$$

$$z'^t_\ell = \text{GPSA}\left(\text{LN}(z^t_{\ell-1})\right) + z^t_{\ell-1}, \quad \ell = 1, \ldots, L$$

$$z^t_\ell = \text{MLP}\left(\text{LN}(z'^t_\ell)\right) + z'^t_\ell, \quad \ell = 1, \ldots, L$$

$$\rightarrow u^t = FE^t([\theta_{task}^t; x_p E]), \text{ where } u^t \text{ is the first element of } z^t_L$$

$$\hat{y}_t = \text{MLP}(u^t)$$

$loss_{clf} = \text{Cross-Entropy}(y_t, \hat{y}_t)$, $y_t$ is the actual target

The $\theta_{task}^t$ learned by the loss function is transferred to the next step.

## 3.2. Feature Distillation

Distillation is the loss that ensures that the model does not forget the information learned in the previous step. The teacher model of distillation is the model of the previous step, $(t-1)$-th task. When the task token of the teacher model is replaced with the $i$-th task token $\theta_{task}^i$, it has already constructed embeddings similar to the $i$-th model. With the same $\theta_{task}^i$, we distil the classification information of the $i$-th task from the teacher model by learning that the feature $FE^{t-1}([\theta_{task}^i; x_p E])$ from the teacher model and $FE^t([\theta_{task}^i; x_p E])$ from the model being trained in the current step have an equal value. The process of distilling information from all the previous steps (rather than only the $i$-th task) is as follows:

$$u_{teacher}^i = FE^{t-1}([\theta_{task}^i; x_p E]) \text{ for } i = 1, \ldots, t-1$$

$$u_{student}^i = FE^t([\theta_{task}^i; x_p E]) \text{ for } i = 1, \ldots, t-1$$

$$loss_{distill} = \sum_{i=1}^{t-1} \|u_{teacher}^i - u_{student}^i\|$$

At this time, the teacher model and task tokens from the previous steps are frozen, and their values do not vary. From these two losses, the total loss used in the entire learning process is calculated as follows:

$$loss_{total} = loss_{clf} + \lambda * loss_{distill}$$

## 4. Experiment

## 4.1. Experiment Setup

CIFAR-100 (Krizhevsky, 2009) is a 32 × 32-pixel color image comprising 100 classes. The training set comprised 500 images per class for 50,000 images. The validation set comprised 100 images per class for 10,000 images. Each dataset was split according to the following four protocols: 1) B0-5steps: 100 classes divided into 5 contexts, with 20 classes per step; 2) B0-10steps: 100 classes divided into 10 contexts, with 10 classes per step; 3) B50-5steps: 50 classes were trained first, and the remaining 50 classes were divided into 5 contexts and trained at 10 classes per step; and 4) B50-10steps: 50 classes were trained first, and the remaining 50 classes were divided into 10 contexts and trained at 5 classes per step. The labels for each task are disjoint. The order of the classes was based on the order of labels in (Douillard et al., 2020, 2022; Huang et al., 2023; Yan et al., 2021). Each step was trained using only the images corresponding to the context unit. The images from the previous steps were not used.

To overcome the low performance compared to the CNN, notwithstanding the use of the attention layer as a GPSA (d'Ascoli et al., 2022), we applied mix-up (Zhang et al., 2018) only in the first step to eliminate performance underestimation, regardless of catastrophic forgetting. We used 12 heads for each self-attention block and stacked six blocks to construct the entire model. The experiments were run on Pytorch1.12. The model was trained using the settings of a single Quadro RTX 8000 GPU.

## 4.2. Comparison with SOTA Methods

The compared algorithms were PASS (Zhu et al., 2021), FeTrIL (Petit et al., 2023), and TCIL (Huang et al., 2023). These are the best-performing models for non-rehearsal-based fields. PASS uses an alternative to the rehearsal memory by extracting prototypes for each class and adding noise. FeTrIL trains only the feature extractor of the first task model and then generates pseudo-features by applying a geometric translation to the data of each task. The pseudo-feature is then trained using a neural network or support vector machine. TCIL uses multi-level distillation (P. Zhou et al., 2020) for logit, features, and rescoring (Zhao et al., 2020) such that the weight norm of the part for new data is on a scale similar to the weight norm of the part for old data. TCIL also uses CNN attention (Woo et al., 2018) to refine and combine features for classification.

We aimed to demonstrate the superiority of our model by comparing it with existing high-performance models. The metrics used for the comparison are as follows:

1) Accuracy: After training with only images in the context of each step, the validation set (including all data from the previous step) was evaluated. In the evaluation, the context information $\mathcal{C}$, indicating the step to which each image belongs, was provided along with the input data $x$; $f_t : x \times \mathcal{C} \to \hat{y}$, and the accuracy was measured using the prediction value $\hat{y}$. The accuracy between the target $y_k$ of the $k$-th step data and the predicted value $\hat{y}_k = f_t(x_k, \mathcal{C})$ after the $t$-th step is denoted by $A_k^t$. The overall accuracy $A^t$ of the $t$-th step was measured as follows (De Lange et al., 2022; van de Ven et al., 2022):

$$A^t = \sum_{k=1}^{t} \alpha_k A_k^t,$$

where $\alpha_k$ is the weight based on the amount of data of the $k$-th step in each step.

2) Backward transfer (BWT): The extent to which a model displays catastrophic forgetting can be assessed by measuring the effect of learning the data in a new step on the representation of the data in the previous step. The overall degree of forgetting at step $t$ is calculated as follows (Chaudhry et al., 2018; Yan et al., 2021):

$$\text{BWT} = \frac{1}{t-1}\sum_{k=2}^{t}\sum_{j=1}^{k} \alpha_k (A_j^k - A_j^j),$$

where $\alpha_k$ is the weight based on the amount of data of the $k$-th step in each step. We measured the metric for each step and analyzed its effect on the previous step.

## 4.3. Experimental Results

We demonstrated that our model outperformed non-rehearsal memory-based methods in terms of accuracy and forgetting. We performed the experiment in a task-IL situation. The model infers labels within the task by identifying the task to which the data belong (De Lange et al., 2022; van de Ven et al., 2022). Accuracy is based on the performance obtained after the model completes training for each step. Fig. 2 shows that the proposed model outperformed the other methods. In the initial tasks, our method started with a marginally lower performance than the other CNN-based methods. As the number of steps increased, our method rapidly surpassed the others in terms of performance, thereby widening the gap. It achieved improvements of 2.97 and 1.99% over the second-best models in the B0-10 and B50-10 step settings, respectively. This tendency was lower for smaller numbers of steps. These results are presented in Supplementary Information. Detailed numerical comparison results are summarized in Table 1.

The extent of forgetting was also assessed. When modifying the weight of the model to fit the new data in the new step, the classification information regarding the data in the previous step may be lost. Therefore, we calculated and compared the BWT at each step as the degree of forgetting by learning new data. The BWT was measured after the second step. See the Methods section for the detailed calculations. The larger the magnitude of the BWT in the negative direction, the higher the forgetting that occurs. As shown in Fig. 3, our method displayed negligible forgetting compared with the other methods. BWT decreased significantly as the number of steps increased. This is unlike TCIL and PASS, which show a significantly large negative increase in BWT as the number of steps increases. It outperformed the second-best model by 0.71% and 1.56% in the B0-10 and B50-10 step scenarios, respectively. The results for a small number of steps are provided in the Supplementary Information. The standard deviation of the BWT also indicates that forgetting does not increase significantly as the number of steps increases.

**Table 1. Performance of the comparative methods under different incremental conditions.** Each scenario indicates the number of classes that were trained initially and the total number of steps. The numbers in the table show the sample means and 95% confidence intervals for the values evaluated on each task's data with the model trained on data from all steps.

| Methods | B0-5steps | | B0-10steps | | B50-5steps | | B50-10steps | |
|---|---|---|---|---|---|---|---|---|
| | Accuracy | BWT (%) | Accuracy | BWT (%) | Accuracy | BWT (%) | Accuracy | BWT (%) |
| PASS | 0.6992 [0.611, 0.7874] | -3.6163 [-4.8634, -2.3691] | 0.7017 [0.6507, 0.7527] | -4.4069 [-5.2397, -3.5742] | 0.7971 [0.7637, 0.8305] | -2.492 [-3.0877, -1.8964] | 0.8321 [0.7405, 0.8621] | -4.2356 [-5.206, -3.2651] |
| FeTrIL | 0.6609 [0.5417, 0.7801] | -0.8596 [-1.34, -0.3792] | 0.6656 [0.5988, 0.7324] | -1.2013 [-1.5225, -0.8801] | 0.8007 [0.7773, 0.8241] | -1.7234 [-1.8611, -1.5556] | 0.8386 [0.797, 0.8793] | -1.9682 [-2.103, -1.8334] |
| TCIL | **0.8547** [0.8198, 0.8896] | -0.5685 [-1.1832, 0.0462] | 0.8394 [0.7534, 0.9254] | -3.3524 [-4.691, -2.0139] | 0.8338 [0.7779, 0.8897] | -1.1144 [-2.0489, -0.1799] | 0.7085 [0.5668, 0.8502] | -7.1538 [-10.5941, -3.7135] |
| **ours** | 0.8363 [0.8104, 0.8622] | **-0.1748** [-0.2386, -0.111] | **0.869** [0.8488, 0.8894] | **-0.4961** [-0.659, -0.3332] | **0.8418** [0.795, 0.8886] | **-0.401** [-0.663, -0.1391] | **0.8585** [0.8094, 0.9076] | **-0.4081** [-0.6457, -0.1704] |

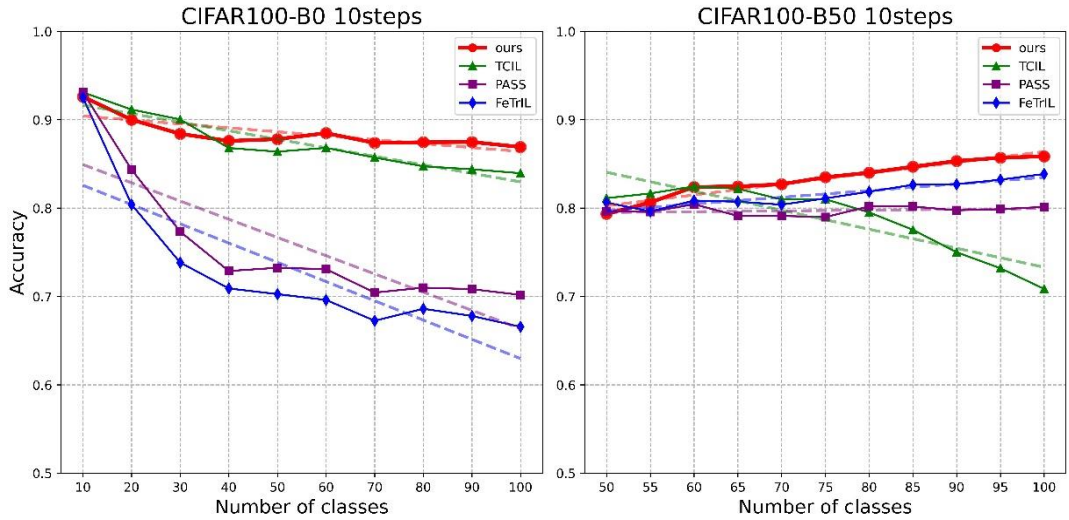

**Fig. 2. Comparative analysis of the performance for each step.** The performances were calculated by averaging the accuracy over the dataset from all steps. The left side was evaluated under the scenario of zero initial classes and 10 incremental steps (= B0-10steps). The right side was evaluated under the scenario of 50 initial classes and 10 incremental steps (= B50-10steps). The dotted lines represent trends.

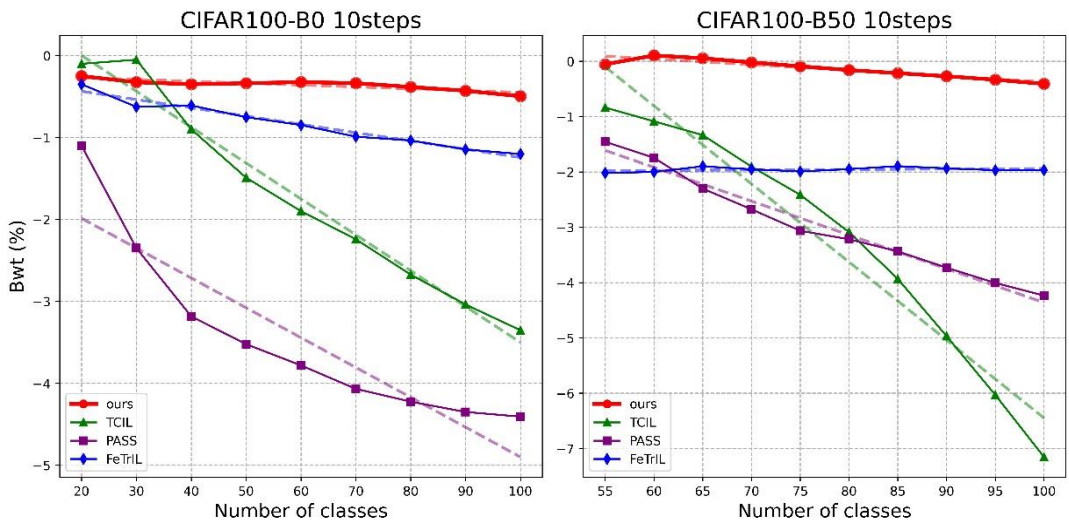

**Fig. 3. Backward transfer (BWT) evaluation for each step.** The BWT is a numerical measure of the effect of the new knowledge gained in each step on the previous knowledge. The left side was evaluated under the scenario of zero initial classes and 10 incremental steps (= B0-10steps). The right side was evaluated under the scenario of 50 initial classes and 10 incremental steps (= B50-10steps). The dotted lines represent trends.

### 4.4. Ablation Study for Task Tokens

To examine whether the task tokens contained sufficient information from all previous steps, we measured the accuracy by combining all task tokens when inferring the data of each task for the final trained model. Fig. 4 shows the results obtained for the B0-10 steps. The highest accuracy was achieved when all task data were combined with the task tokens of the corresponding task ID. As the task progresses from one step to the next, the task token from the previous step is inherited and transformed into a state containing the information for that step. Consequently, as the steps progressed, the task token tended to forget information from the earlier steps. The difference in accuracy between evaluating the data using the correct task token and those using the immediate subsequent step task token was 9.8% on average, with a standard deviation of 5.32%. The average difference in accuracy between the correct and least effective task tokens was 68.4%, with a standard deviation of 11.17%. When the correct token was used, the average accuracy across all 10 tasks was 86.91% with a standard deviation of 3.11%.

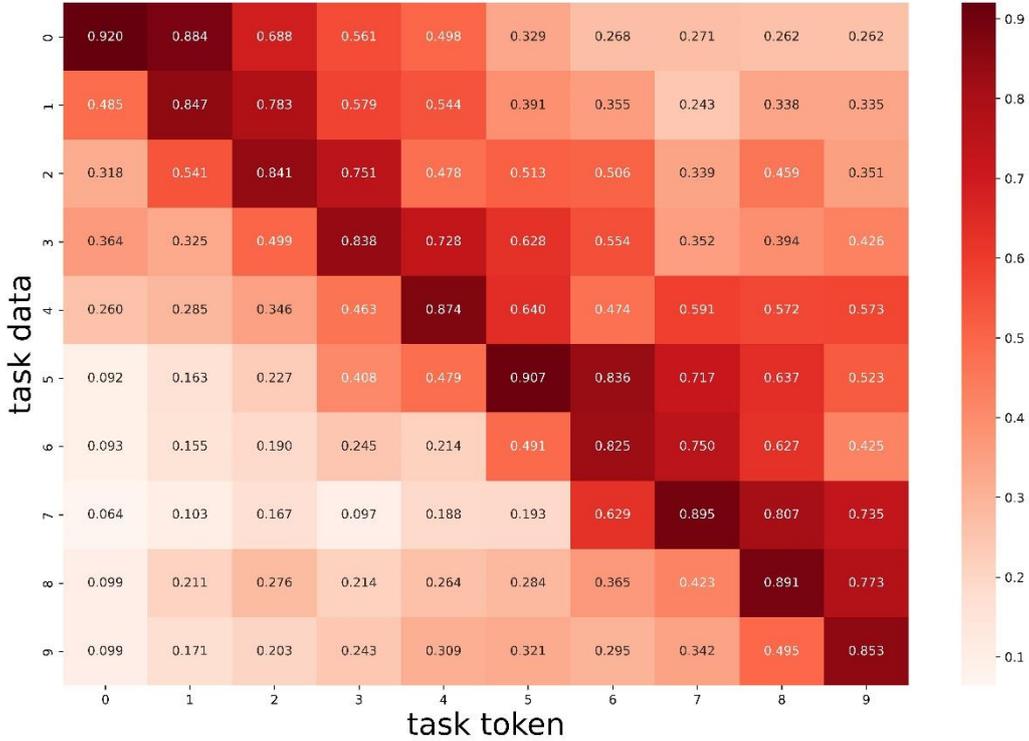

**Fig. 4. Effectiveness of task token with the most appropriate token and data pairs.** Accuracy comparison for each pair of data and task token for each task for a model that was fully trained till the final step of the B0-10steps scenario. The x-axis indicates the task token used, and the y-axis indicates the task's data used. For example, the point (5,2) represents the accuracy obtained using the fifth task token to evaluate the data from the second task.

## 5. Discussion

In this study, we introduced a novel approach for incremental learning that consolidates new knowledge while preserving existing information. This effectively mitigated the issue of catastrophic forgetting without relying on rehearsal memories. Our method outperformed other non-rehearsal memory-based methods in task-incremental learning scenarios. This was demonstrated through comprehensive experiments conducted in various learning settings. Furthermore, by adopting a strategy inspired by the ViT class token (Dosovitskiy et al., 2021), we expanded the potential for incremental learning within the ViT model family. Finally, a notable aspect of our research is the proposal of a structure that enables the selective forgetting and retention of specific knowledge. This comprehensive approach presents a robust solution for incremental learning that effectively addresses key security challenges associated with storing past data.

Distillation was performed for each task token to prevent catastrophic forgetting. As there was no rehearsal memory, the data for the labels learned in the previous task did not exist. In this situation, it is difficult for general distillation (Hinton et al., 2015), which receives knowledge through the Kullback–Leibler divergence with a logit from the teacher model, to ensure good performance. Therefore, distillation is performed only until the feature extraction step (Zhu et al., 2021). Each classifier of past tasks does not permit a gradient flow after learning that step. Therefore, the patient's body weight remained unaltered. Consequently, the continuity of knowledge is maintained based on the similarity between the features input into these classifiers and those on which the classifiers were trained in the previous steps. Although distillation has been used in several studies to prevent forgetting (Douillard et al., 2020, 2022; Wang et al., 2022; P. Zhou et al., 2020; Zhu et al., 2021), determining the weight that preserves both old and new knowledge without rehearsal memory is difficult. Therefore, we used a ViT structure with an attention mechanism (rather than a CNN) to embed features in different directions even within the same model weight. In addition to the patch tokens obtained from each image, the attention block of ViT contains a class token shared by all batches. Moreover, knowledge regarding classification accumulates during the embedding of class tokens. Utilizing this characteristic, we used a task token to provide specific attention to each task and accumulate knowledge. In DyTox (Douillard et al., 2022), a rehearsal-memory-based task token is placed in the middle to form an encoder/decoder structure. However, because we do not have rehearsal memory, the task token should contain model-wide information. Therefore, the task token was attached immediately after patch tokenization. Multiple task tokens for an individual model have an effect identical to running multiple models for different tasks. Therefore, distilling the student model from the teacher model to produce similar features for each task token has an effect identical to that of distilling from multiple teachers separately generated for each task. Based on this characteristic, it is feasible to select the knowledge to be remembered or forgotten without using a specific task token. As each task has its own distinctive task token and classifier, removing a specific task token or classifier effectively eliminates the knowledge of that step. In addition, because we presented a learning methodology that uses class tokens, it has the advantage of being scalable in the ViT series of models (d'Ascoli et al., 2022; Dosovitskiy et al., 2021; Hassani et al., 2022; Liu et al., 2021; Touvron et al., 2021; Wu et al., 2021).

In most scenarios, our model is more robust to catastrophic forgetting than other methods. For B0-5steps, we observed that the accuracy of TCIL was higher by 1.84%p. This is because TCIL forgets old tasks more as the number of steps increases (see Supplementary Fig. 1). For the B0-10steps, we determined that TCIL's classification performance reduces to 0.4 for data from the first task. This causes a reduction in overall performance. However, we did not observe this trend in 5steps. As the threshold step for forgetting is not within 5steps, it appears that the TCIL model performs marginally better in this scenario. In a scenario where 50 classes are pre-trained, if we consider pre-training as part of the task, the measure of overall accuracy would be significantly influenced by the accuracy of the first task of the 50 classes. Therefore, TCIL performs marginally poorer than our model in the B50-5steps scenario because of the marginal forgetting of the initial task. In the B50-10 step scenario, the accuracy decreased abruptly in the middle of the total number of steps. This trend was also observed in the BWT. Here, our model outperformed the BWT even in the B0-5steps setting with a higher TCIL in the final performance.

This model was optimized for incremental task learning using task information. However, it can also be used in class-incremental learning models. This can be inferred without task information using all tokens from the first to the final task to obtain logit values for all the task labels, concatenate all the logits, and identify the index with the largest value (the training process is identical). However, this model did not include a loss term to ensure that each task was effectively discriminated from the others (Yan et al., 2021). Therefore, it relies on learning sufficiently large logit values for the correct labels for each task. In our experiments, we used a mix-up (Zhang et al., 2018) as the first step (a weighted sum of the data from the two images) and mixed the labels of the two images in equal proportions. Although this method improves the performance, it is inappropriate for class-incremental learning. This is because it was trained such that the logit value of the correct label was not sufficiently large. Therefore, they cannot be clearly distinguished from the logits generated by the classifiers of other tasks. We performed the experiment with the B0-10 scenario without mix-up in the first step and obtained a final accuracy of 0.857 for task-IL and 0.45 for class-IL (see Supplementary Fig. 3). These values are higher than the final values reported for class-IL in a TCIL study. This indicates that the methodology can be extended to class-ILs to achieve good performance. Furthermore, the development of a task-distinguishable module similar to the auxiliary classifier in the DER (Yan et al., 2021) can potentially refine and advance this extension.

# 6. Conclusion

In this study, we propose a model that effectively preserves existing knowledge and seamlessly integrates new information. This circumvents the need to re-access the previous datasets. Distillation using task tokens has an effect similar to that of running multiple teacher models simultaneously. It is scalable to any model of the ViT family (d'Ascoli et al., 2022; Dosovitskiy et al., 2021; Hassani et al., 2022; Liu et al., 2021; Wu et al., 2021). Our approach surpasses prior non-rehearsal memory models by providing enhanced classification accuracy in a task-incremental learning setting and by significantly reducing the rate of forgetting. We also calculated the combination of data from all tasks and the classification performance based on task tokens to verify that task tokens effectively contained information relevant to classification. However, our study has some limitations that should be addressed in future studies. Using knowledge distillation with all task tokens for each incremental task increases the computational cost because training involves all tasks as the number of tasks increases. Future work should explore ways to reduce the computational costs. Training is performed with all task tokens; however, there are methods to select only a few appropriate tokens for training.

## CRediT authorship contribution statement

**Young Jo Choi**: Conceptualization, Methodology, Visualization, Writing of the original draft. **Min Kyoon Yoo**: Writing, reviewing, and editing. **Yu Rang Park**: Conceptualization, Investigation, Supervision.

## Declaration of competing interest

The authors declare that they have no known competing financial interests or personal relationships that could have appeared to influence the work reported in this paper.

## Code availability

The source code and pipeline to reproduce our study can be accessed at
https://github.com/DigitalHealthcareLab/23CLwithoutRM

## Acknowledgement

This work was supported by the Bio-Industrial Technology Development Program (20014841), funded By the Ministry of Trade, Industry & Energy (MOTIE, Korea).